# HyCIR: Boosting Zero-Shot Composed Image Retrieval with Synthetic Labels

**Yingying Jiang, Hanchao Jia, Xiaobing Wang, Peng Hao**

Samsung R&D Institute China-Beijing

{yy.jiang, hanchao.jia, x0106.wang, peng1.hao}@samsung.com

## Abstract

Composed Image Retrieval (CIR) aims to retrieve images based on a query image with text. Current Zero-Shot CIR (ZS-CIR) methods try to solve CIR tasks without using expensive triplet-labeled training datasets. However, the gap between ZS-CIR and triplet-supervised CIR is still large. In this work, we propose Hybrid CIR (HyCIR), which uses synthetic labels to boost the performance of ZS-CIR. A new label Synthesis pipeline for CIR (SynCir) is proposed, in which only unlabeled images are required. First, image pairs are extracted based on visual similarity. Second, query text is generated for each image pair based on vision-language model and LLM. Third, the data is further filtered in language space based on semantic similarity. To improve ZS-CIR performance, we propose a hybrid training strategy to work with both ZS-CIR supervision and synthetic CIR triplets. Two kinds of contrastive learning are adopted. One is to use large-scale unlabeled image dataset to learn an image-to-text mapping with good generalization. The other is to use synthetic CIR triplets to learn a better mapping for CIR tasks. Our approach achieves SOTA zero-shot performance on the common CIR benchmarks: CIRR and CIRCO.

## Introduction

Image retrieval plays an important role in many tasks, e.g., product retrieval, landmark retrieval, person re-identification, and gallery photo search. Composed image retrieval (CIR) allows searching images with both a query image and a query text. In CIR, the user's retrieval intention can be more precisely represented with both image and text.

Recently, CIR is a very hot research topic. Several methods (Baldrati et al. 2022, Jang et al. 2024, Xu et al. 2024) are proposed to support CIR based on strong triplet supervision from the CIR dataset. However, as the labeling of CIR triplets are expensive and time-consuming, current CIR datasets are small. In these methods, one model is often trained for each dataset and the generalization is not good. Zero-shot CIR (ZS-CIR) is proposed to support various CIR tasks without the need of CIR dataset. Some approaches (Saito et al. 2023, Liu et al. 2023, Baldrati et al. 2023, Tang et al. 2024, Gu et al. 2024) utilize a frozen visual and text encoder and train an image-to-text mapping network to represent im-age as a pseudo-word token. Their training requires only unlabeled images, or image-caption pairs, or even text dataset. Besides this, training-free approaches (Karthik et al. 2024) are also explored for ZS-CIR. Off-the-shelf vision-language model and LLM are used to generate a description of the desired target image and perform text-to-image retrieval. These ZS-CIR methods tend to have good generalization as they often rely on models trained with large-scale data. However, due to the lack of effective triplet supervision, the gap between ZS-CIR and triplet-supervised CIR is still large.

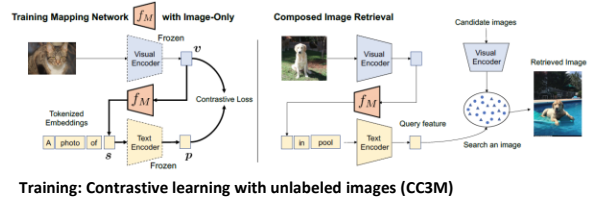

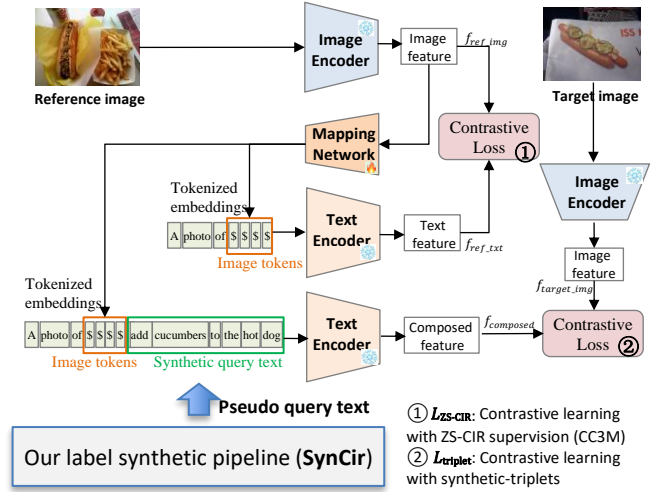

Figure 1: Overview of our solution HyCIR. (a) Training and inference of Pic2Word (Saito et al. 2023). (b) Our hybrid training with synthetic triplets and large-scale unlabeled images. It combines contrastive learning with ZS-CIR supervision and contrastive learning with synthetic triplets obtained by our label synthesis pipeline SynCir. The inference is similar to Pic2Word.

| Dataset | Data labeling approach | Required resources | Label diversity | Scalability |
|---|---|---|---|---|
| **CIRR** (Liu et al., 2021) | Human annotation | Images | Yes | Hard |
| **Laion-CIR** (Liu et al. 2023) | Retrieval-based pipeline to construct dataset | Large-scale image-text pairs | Yes | Easy |
| **LaSCo** (Levy et al., 2024) | Generate labels from VQA2.0 samples by GPT-3 | VQA samples | Yes | Medium |
| **VDG** (Jang et al., 2024) | Generate labels by tuned LLM-based visual delta generator | CIR triplets for VDG tuning | Yes | Easy |
| **SPN** (Feng et al., 2024) | Generate labels by MLLM and prompt templates | Images | No | Easy |
| **SynCir (Ours)** | Generate labels by image caption and LLM and filter data | Images | Yes | Easy |

Table 1: Data creation methods for Composed Image Retrieval task. Our method generates diverse labels from unlabeled images without additional information, which make it easy to scale up to more data.

We propose Hybrid CIR (HyCIR), in which the key idea is to use synthetic triplets to boost the performance of ZS-CIR (Figure 1). As triplet supervision can be used to learn the CIR task well from <reference image, query text, target image> triplets, we can generate pseudo triplet labels and make them work together with current ZS-CIR methods to improve the performance of ZS-CIR. In specific, we propose a new pipeline SynCir to generate synthetic labels for CIR, in which only unlabeled images are required. The pipeline includes three steps: 1) visual similarity-based image pair extraction, 2) vision-language model and LLM based query text generation, and 3) language similarity-based data filter. To make our synthetic dataset work together with existing ZS-CIR methods, we propose a hybrid training strategy. Two kinds of contrastive learning are adopted. One is to use unlabeled image dataset to learn an image-to-text mapping with good generalization. The other is to use synthetic CIR triplets to learn a better mapping for CIR tasks. In our experiments, we use unlabeled images from COCO (Lin et al. 2014) train dataset to generate synthetic labels and utilize Pic2Word (Saito et al. 2023) as the baseline ZS-CIR method. Results show that we can achieve SOTA performance on CIRR (Liu et al. 2021) and CIRCO (Li et al. 2024). In addition, although we don't generate pseudo labels for fashion-style images, we can still have better performance than the baseline on FashionIQ (Guo et al. 2019). Our contributions are as follows:

- We propose Hybrid CIR (HyCIR), that uses synthetic labels to boost the performance of ZS-CIR with a hybrid training strategy.
- A new pipeline SynCir is proposed to generate labels for CIR, which consists of image pair extraction, label generation, and data filter. With SynCir, we can generate diverse labels from unlabeled images, which makes it easy to scale up to more data.
- We introduce a hybrid training strategy to ZS-CIR, which combines contrastive learning for ZS-CIR with large-scale unlabeled images and contrastive learning with synthetic CIR triplets.
- Our experiments show that the proposed solution can achieve SOTA zero-shot performance on CIRR test set (R@5: 69.03%) and CIRCO test set (mAP@5: 18.91%).

## Related Work

### Zero-Shot Composed Image Retrieval

Zero-shot composed image retrieval (ZS-CIR) has been proposed and studied recently. Some approaches utilize token-based methods (Saito et al. 2023, Baldrati et al. 2023, Gu et al. 2024, Tang et al. 2024, Du et al. 2024, Li et al. 2024, Suo et al. 2024). They use a frozen visual and text encoder and train an image-to-text mapping network to represent image as a pseudo-word token and then search with the pseudo-word token and query text. Pic2Word (Saito et al. 2023) and SEARLE (Baldrati et al. 2023) are such ZS-CIR methods and their image-to-text mapping network are trained only with unlabeled images. LinCIR (Gu et al. 2024) adopts another strategy that relies solely on language for training, which leverages self-supervision through self-masking projection. Context-I2W (Tang et al. 2024), ISA (Du et al. 2024) and KEDs (Suo et al. 2024) try to enhance the image representation. Context-I2W tries to train an Intent View Selector and Visual Target Extractor to focus on partial part of image rather than the whole image. ISA utilizes an adaptive token learner to map the visual feature map to a series of sentence tokens instead of only one word token. KEDs propose to augment the pseudo-word token with external knowledge. MCL (Li et al. 2024) further improves mapping or aligning of the vision and language input based on multimodal composition learning. Besides token-based ZS-CIR methods, linear interpolation based ZS-CIR method (Jang et al. 2024) is proposed, which directly combines image and text embeddings to produce a composed embedding.

Training free ZS-CIR methods are also explored. CI-ReVL (Karthik et al. 2024) uses off-the-shelf vision-language model and LLM to generate a description of the desired target image and perform text-to-image retrieval.

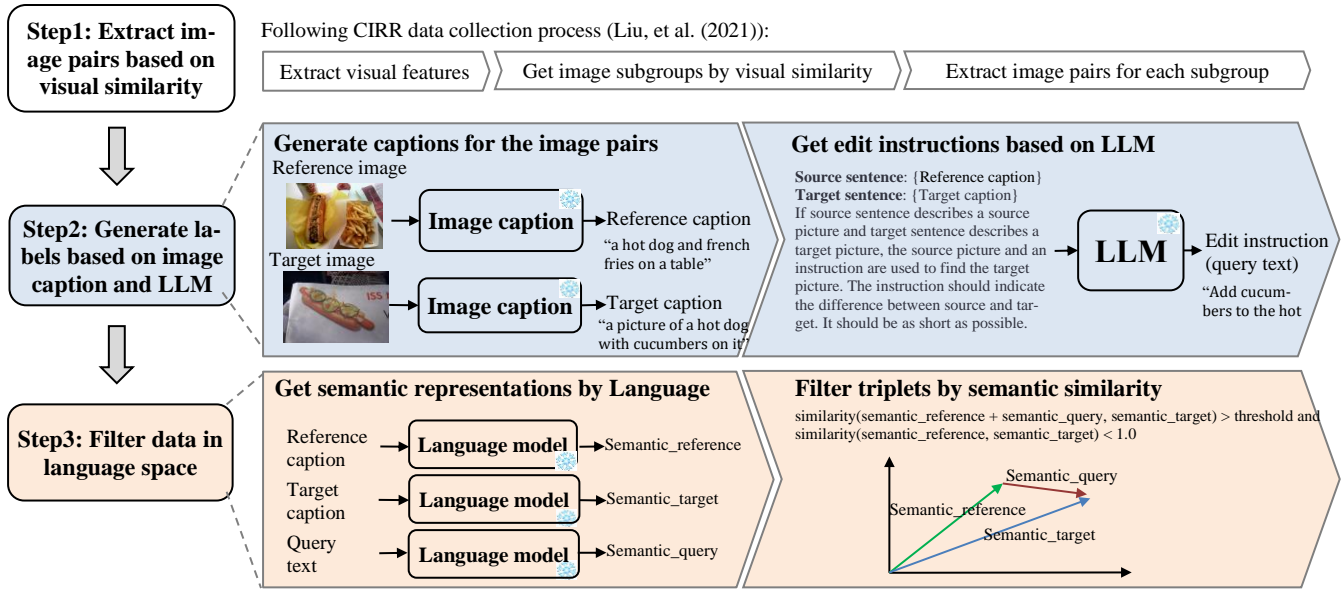

Figure 2: Our vision-language model and LLM based label synthesis pipeline SynCir for CIR

TFCIR (Sun et al. 2023) is also based on off-the-shelf models and it transforms implicit interaction between query text and image into explicit local text using a local concept re-ranking.

In our work, we introduce triplet supervision from synthetic CIR dataset to ZS-CIR, and the proposed HyCIR can boost the performance of ZS-CIR significantly.

## Label Synthesis for CIR

Pseudo triplet generation for CIR has been studied by researchers. Table 1 shows some methods. LaSCo (Levy et al. 2024) is a synthetic dataset constructed by generating query texts from VQA2.0 samples by GPT-3. VQA data is required in the label synthesis process. VDG (Jang et al. 2024) uses tuned LLM-based Visual Delta Generator to generate pseudo triplets. Triplet supervision, such as CIRR (Liu et al. 2021), is needed to get the VDG. SPN (Feng, et al., 2024) leverages MLLM and prompt templates to generate labels. The templates may limit the supported input style of query text for CIR. In the work of ZS-CIR (Liu et al. 2023), they propose to construct a triplet dataset automatically for training ZS-CIR model by exploiting a large-scale dataset of image-text pairs, Laion-COCO. The quality of the generated label may be influenced by the text-based image retrieval performance. Our label synthesis method is different from these methods that our triplets are generated with only unlabeled images, it is based on pretrained vision-language models and LLM and is training-free, it can generate labels with diversity based on LLM, and it can easily scale up to more data.

## Approach

Figure 1 shows the overview of our solution HyCIR. We propose to use synthetic data to enhance the performance of zero-shot composed image retrieval (ZS-CIR). We first generate pseudo triplets for CIR with our vision-language model and LLM based label synthesis pipeline SynCir. Then we perform hybrid training to utilize both ZS-CIR supervision and synthetic triplet supervision to learn a better image-to-text mapping, which can represent the reference image to several image tokens. During the inference, image and text feature are used to get the composed query feature to find the target image from the candidate images.

## Pseudo Triplet Generation for CIR

The task of CIR is to find the target image given the reference image and query text. The CIR datasets often consist of triplets of <reference image, query text, target image>. To generate triplets automatically from unlabeled images, we need to find image pairs of <reference image, target image> first and then generate the query text for each image pair. After that, the data can be filtered to remove those samples that are not good enough. Figure 2 illustrate our CIR label synthesis pipeline SynCir, which is based on pre-trained vision-language model and LLM.

### Image pair extraction

To extract image pairs, we follow the process of CIRR (Liu et al. 2021). We first extract visual features by resnet152 for each image in the unlabeled image dataset. We then extract subgroups based on the similarity between image features. Each subgroup has 6 images. The distance between the first image and subsequent image is less than 0.94 and the distance between subsequent images should be larger than 0.002. For each subgroup, we extract 9 image pairs. At last, we remove repeated image pairs. In this way, the reference image and target image in each image pair are not very similar, and the image pairs are also different.

**Query text generation**
When we get image pairs, we generate query text for each of them. Given an image pair <reference image, target image>, we can generate the reference caption and target caption with a pretrained vision-language model. And we then leverage a pretrained LLM model to generate the edit instruction (query text). The prompt for the LLM describes the CIR task and ask the LLM to generate edit instruction (i.e., query text) given the reference caption and target caption. The prompt is shown in Figure 3.

Due to the powerful capability of LLM, the query text for each image pair can be generated automatically.

> **Source sentence**: {Reference caption}
> **Target sentence**: {Target caption}
> If source sentence describes a source picture and target sentence describes a target picture, the source picture and an instruction are used to find the target picture. The instruction should indicate the difference between source and target. It should be as short as possible. Show the instruction.

Figure 3: Prompt used by LLM to generate query text

**Data filter in language space**
We further filter the generated data in language space. Two conditions are considered: 1) the similarity between the reference caption and target caption; 2) the quality of the generated query text. We use a language model to get the semantic embeddings for the reference caption, query text and target caption. They are represented as *semantic_reference, semantic_query,* and *semantic_target*. We use the similarity between embeddings to measure the two conditions.

On the one hand, if the reference caption and target caption are the same, we remove the sample from the dataset, because either the images in the image pair are very similar or the generated caption does not include the descriptions of different contents between the reference image and target image. On the other hand, if the generated query text is not consistent with the reference caption and target caption, we also remove the sample. In specific, when the similarity between *semantic_reference+semantic_query* and *semantic_target* is smaller than a threshold, we consider the generated query text is not consistent with the reference caption and target caption and the sample will be dropped.

## Hybrid ZS-CIR Training

We design a hybrid training strategy for ZS-CIR. As shown in Figure 1, we adopt a pseudo-word token based ZS-CIR method as the baseline and extend it to train with both large scale unlabeled image data and synthetic triplets.

**Model architecture**
We use Pic2Word (Saito et al. 2023) as the ZS-CIR baseline. It utilizes a frozen visual and text encoder and train an image-to-text mapping network to represent image as a pseudo-word token. Recent work (Xu et al. 2024, Du et al. 2024) adopts multiple pseudo tokens instead of one token to improve the image representation. We also adopt multiple word tokens to represent the image.

Given a reference image and query text, the retrieval can be performed by comparing the composed feature $f_{composed}$ with image feature of candidate images. The mapping network maps the reference image into image tokens, represented by [$ $ $ $]. The query can be composed by a photo of [$ $ $ $], [query text]. And the text encoder then encodes the query to get the composed feature $f_{composed}$ for image retrieval.

**Hybrid training**
Following Pic2Word, we only train the mapping network to map an image to multiple tokens. As we use both unlabeled large scale image data and synthetic triplets for training, we use two contrastive learning separately. To keep the generalization ability of ZS-CIR, we use a contrastive loss to learn from large-scale unlabeled images, which is similar to Pic2Word. To improve the performance of ZS-CIR, we use another contrastive loss to learn from synthetic triplets.

Suppose M is the parameters of the mapping network, we propose to minimize the hybrid loss $L_{hybrid}$ with respect to the mapping network.

$$\min_{M} L_{hybrid} = L_{ZS-CIR} + L_{triplet} \quad (1)$$

which includes two terms:

$$L_{ZS-CIR} = L_{contrastive}(f_{ref\_img}, f_{ref\_txt}) \quad (2)$$
$$L_{triplet} = L_{contrastive}(f_{composed}, f_{target\_img}) \quad (3)$$

where $L_{ZS-CIR}$ is the contrastive loss between reference image feature $f_{ref\_img}$ and its corresponding text feature $f_{ref\_txt}$, and $L_{triplet}$ is the contrastive loss between the composed feature $f_{composed}$ and target image feature $f_{target\_img}$.

# Experiments

We evaluate our method on the standard CIR benchmarks. After introducing experimental setup, we compare our results with SOTA ZS-CIR methods, and then introduce our ablation study on the dataset, model, training strategy and generalization ability. Lastly, we discuss about our findings and future work.

| Zero-shot CIR methods | Training data | CIRR test | | | | | | | CIRCO test (mAP@K) | | | |
|---|---|---|---|---|---|---|---|---|---|---|---|---|
| | | R@1 | R@5 | R@10 | R@50 | Rs@1 | Rs@2 | Rs@3 | K=5 | K=10 | K=25 | K=50 |
| SEARLE (Baldrati et al. 2023) | ImageNet test | 24.24 | 52.48 | 66.29 | 88.84 | 53.76 | 75.01 | 88.19 | 11.68 | 12.73 | 14.33 | 15.12 |
| ISA (Du et al. 2024) | CC3M | 30.84 | 61.06 | 73.57 | 92.43 | - | - | - | 11.33 | 12.25 | 13.42 | 13.97 |
| CASE (Levy et al. 2024) | LasCo.Ca. | 35.40 | 65.78 | 78.53 | 94.63 | 64.29 | 82.66 | 91.61 | - | - | - | - |
| CoVR (Lucas et al. 2023) | WebVid | **38.48** | 66.70 | 77.25 | 91.47 | **69.28** | 83.76 | 91.11 | - | - | - | - |
| MCL (Li, et al. 2024) | MMC | 26.22 | 56.84 | - | 91.35 | | | | 17.67 | 18.86 | 20.80 | 21.68 |
| Slerp + TAT (Jang, et al. 2024) | CC3M+LLaVAlign+Laion-2B | 33.98 | 61.74 | 72.70 | 88.94 | 68.55 | **85.11** | **93.21** | 18.46 | 19.41 | 21.43 | 22.41 |
| Pic2Word-CLIP (Saito et al. 2023) | CC3M | 23.90 | 51.70 | 65.30 | 87.80 | 53.63 | 74.36 | 87.27 | 8.72 | 9.51 | 10.64 | 11.29 |
| Pic2Word-BLIP | CC3M | 33.1 | 63.83 | 74.98 | 92.33 | 63.08 | 82.12 | 91.73 | 10.71 | 11.34 | 12.49 | 13.18 |
| **Pic2Word-CLIP-HyCIR** | CC3M+synthetic data | 25.08 | 53.49 | 67.03 | 89.85 | 53.83 | 75.06 | 87.18 | 14.12 | 15.02 | 16.72 | 17.56 |
| **Pic2word-BLIP-HyCIR** | CC3M+synthetic data | 38.28 | **69.03** | **79.71** | **95.27** | 66.79 | 84.79 | 93.06 | **18.91** | **19.67** | 21.58 | **22.49** |

Table 2: Comparison with SOTA zero-shot composed image retrieval methods on CIRR test and CIRCO test. The best scores are marked in bold, while the second best are underlined. Pic2Word-CLIP: baseline method; Pic2Word-BLIP: Pic2Word with BLIP encoder; Pic2Word-CLIP-HyCIR: with CLIP encoder and use hybrid training with CC3M and our synthetic dataset; Pic2Word-BLIP-HyCIR: with BLIP encoder and hybrid training.

## Experimental Setup

### Evaluation datasets

**CIRR** dataset (Liu et al., 2021) contains 21,552 real-world images and includes 36,554 triplets in total, divided into 3 subsets with 80% in training, 10% in validation, and 10% in testing. Specifically, it is configured with 4,351 subgroups, each containing six similar images. We report the CIR results with recall scores at top K retrieval results (R@K) and results under collected subsets ($R_s$@K). **CIRCO** (Li et al. 2024) is an open-domain zero-shot CIR benchmark with real-world images and multiple annotated ground truths. It comprises a total of 1020 queries, randomly divided into 220 and 800 for the validation and test set. Following the metric used in the benchmark (Li et al. 2024), we use the mean Average Precision (mAP@K) for evaluation. **FashionIQ** (Guo et al. 2019) includes 77,684 fashion images categorized into three groups (Dress, Toptee, and Shirt) and organized into triplets.

### Implementation Details

In our implementation of SynCir, BLIP (Li et al. 2022) is used for image caption, GPT-3.5 is utilized as LLM to generate query texts, and Bert (Devlin et al. 2018) is adopted to get the semantic embedding for the data filter. The threshold to judge the similarity between *semantic_reference+semantic_query* and *semantic_target* is set to 0.7. We generate triplets on 82,556 unlabeled real-life images from COCO (Lin et al. 2014) train dataset. 680k triplets is generated based on vision-language model and LLM. After data filter, 650k triplets are reserved.

To verify hybrid training, we use unlabeled images from CC3M (Sharma et al. 2018) to minimize $L1_{ZS-CIR}$ and the synthetic triplets obtained by SynCir to minimize $L2_{triplet}$. In our experiments, the learning rate is set to 1e-4. Adam optimizer is used. In each batch, we use 512 samples from CC3M and 256 samples from synthetic CIR dataset to train the mapping network.

As for the model, we conduct experiments on two visual and text encoders. One encoder is ViT-L/14 CLIP (Radford et al. 2021) pretrained on 400M image-text paired data. The other encoder is ViT-B/16 BLIP (Li et al. 2022) which is finetuned on COCO for Image-Text Retrieval. Regarding token number in the image to text mapping network, we choose to use 4 tokens in our final model. As our synthetic triplets are related to real-world images, we benchmark with SOTA ZS-CIR methods on CIRR test set and CIRCO test set, which are also real-world images. We conduct ablation study on CIRR val set and we further verify the generalization ability on FashionIQ, which consists of fashion images that are different from real-world images.

## Comparison with State-of-the-art Methods

Table 2 shows our results on CIRR test set and CIRCO test set. Our solution achieves higher performance than other ZS-CIR methods. For example, our BLIP-based solution Pic2word-BLIP-HyCIR achieves a 69.03% R@5 on CIRR test set and a 18.91% mAP@5 on CIRCO test set.

As for the training strategy, the proposed hybrid training approach with large-scale unlabeled images from CC3M and synthetic triplets obtained by SynCir can improve the performance of ZS-CIR methods. For CLIP-based method, the hybrid training brings a noticeable performance improvement over the baseline Pic2Word method. For example, it brings a 1.79% increase in R@5 between Pic2Word and Pic2Word-CLIP-HyCIR on CIRR test set, and a 5.4% increase in mAP@5 on CIRCO test set. For BLIP-based method, the hybrid training leads to even higher performance improvement, with a 5.2% increase in R@5 on CIRR test set and a 8.2% increase in mAP@5 on CIRCO test set.

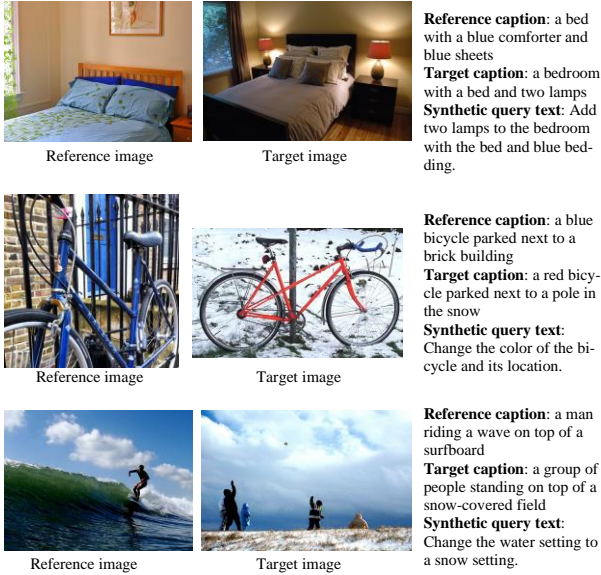

Figure 4: Examples of generated labels by SynCir pipeline

Moreover, as for the visual and text encoder, the BLIP-based method can achieve higher performance than the CLIP-based method.

## Ablation Studies

We conduct ablation studies on CIRR val set to verify the effectiveness of our synthetic dataset, the influence of token number in mapping network on CIR performance and different training strategies, and then evaluate the generalization ability of our solution on FashionIQ.

### Synthetic dataset on ZS-CIR

Figure 4 shows some examples of labels generated by our SynCir pipeline. It can be seen that the proposed SynCir can generate triplets well. Based on two captions of reference image and target image, LLM can generate the query text based on the prompt illustrated in Figure 3.

While our synthetic dataset is based on COCO train images, LaSCo (Levy et al. 2024) is also a synthetic dataset based on COCO VQA samples. We compare the effectiveness of LaSCo and our synthetic dataset generated by SynCir on CIRR val set. In this experiment, we use BLIP based visual and text encoder and utilize 1 pseudo token to represent image. As shown in Table 3, our vision-language model and LLM based label synthesis method is better than the VQA based synthesis method (average recall on CIRR val: 70.43% vs. 69.99%). And our data filter mechanism can further improve the average recall to 70.73%. In the following experiments, we use the filtered synthetic dataset as the synthetic dataset.

| Training data | R@1 | R@5 | R@10 | R@50 | Avg |
|---|---|---|---|---|---|
| CC3M + LaSCO | 37.04 | 68.33 | 79.83 | 94.78 | 69.99 |
| CC3M + synthetic data by SynCir (w/o filter) | **37.50** | 68.93 | 80.45 | 94.85 | 70.43 |
| CC3M + synthetic data by SynCir (w/ filter) | 37.45 | **69.52** | **80.96** | **95.00** | **70.73** |

Table 3: Results on CIRR val with different training data

### Token number in mapping network

As we utilize multiple tokens to represent an image in the image-to-text mapping network, we verify the influence of token number on the CIR performance. We perform the experiment with BLIP visual and text encoder. The models with different token numbers are trained in a hybrid training way with CC3M and synthetic dataset. Table 4 shows the results on CIRR val set. We find we can represent an image with 4 tokens to get a relatively better performance (average recall on CIRR val: 72.14%), and 4 tokens is utilized in the following experiments.

| Token num | R@1 | R@5 | R@10 | R@50 | Avg |
|---|---|---|---|---|---|
| 1 | 37.45 | 69.52 | 80.96 | 95.00 | 70.73 |
| 2 | 38.05 | 70.62 | 81.46 | 95.26 | 71.34 |
| 3 | 39.08 | 71.17 | 82.01 | 95.07 | 71.83 |
| 4 | **39.22** | 71.56 | **82.30** | **95.50** | **72.14** |
| 5 | 39.17 | 71.34 | 82.15 | 95.19 | 71.96 |
| 6 | 38.50 | 71.63 | 81.89 | 95.31 | 71.83 |
| 7 | 38.53 | 70.77 | 82.27 | 95.47 | 71.76 |
| 8 | 38.74 | **71.99** | 81.89 | 94.73 | 71.83 |

Table 4: Results on CIRR val with different token numbers

### Training strategy

The proposed HyCIR adopts a hybrid training strategy. As shown in Formula 1, the hybrid training loss $L_{hybrid}$ consists of two parts: $L_{ZS-CIR}$ and $L_{triplet}$. We compare different strategies to use them on both CLIP-based method and BLIP-based method. Table 5 shows the results on CIRR val set. We can see that the hybrid training $L_{hybrid}$ can achieve better performance than just using one contrastive loss $L_{ZS-CIR}$ or $L_{triplet}$. Moreover, we find the triplet based contrastive learning with synthetic data generated by SynCir can get better performance than the ZS-CIR contrastive learning with CC3M.

| Encoder | Training loss | R@1 | R@5 | R@10 | R@50 | Avg |
|---|---|---|---|---|---|---|
| CLIP | $L_{ZS-CIR}$ | 22.54 | 51.69 | 64.76 | 86.19 | 56.29 |
| | $L_{triplet}$ | 24.56 | 52.47 | 66.73 | 86.55 | 57.57 |
| | $L_{hybrid}$ | **25.59** | **55.8** | **69.14** | **89.93** | **60.11** |
| BLIP | $L_{ZS-CIR}$ | 34.29 | 65.74 | 77.42 | 93.8 | 67.81 |
| | $L_{triplet}$ | 38.81 | 70.96 | 82.01 | 95.07 | 71.71 |
| | $L_{hybrid}$ | **39.22** | **71.56** | **82.3** | **95.5** | **72.14** |

Table 5: Comparison of training strategies on CIRR val.

**ZS-CIR on FashionIQ**
We verify the generalization ability of HyCIR on FashionIQ, which consists of fashion images that are different from real-world images used in our synthetic dataset. The results are shown in Table 6. We can see that although the data distribution of our synthetic dataset obtained by SynCir is quite different from fashion data in FashionIQ, our hybrid training can still achieve better performance than the baseline method. The generalization ability of proposed solution is fine.

| Encoder | Dataset | Dress | | Shirt | | Toptee | |
|---|---|---|---|---|---|---|---|
| | | R@10 | R@50 | R@10 | R@50 | R@10 | R@50 |
| CLIP | CC3M | **20** | 40.2 | 26.2 | 43.6 | 27.9 | 47.4 |
| | CC3M+ synthetic data | 19.98 | **40.8** | **27.62** | **44.94** | **28.14** | **47.67** |
| BLIP | CC3M | 13.28 | 27.26 | 17.95 | 34 | 18.25 | 34.98 |
| | CC3M+ synthetic data | **18.88** | **34.5** | **22.52** | **37.58** | **22.13** | **40.33** |

Table 6: ZS-CIR Results on FashionIQ val

## Discussion

**Label synthesis pipeline. (a) It is easy to scale up the synthetic dataset.** As our pipeline SynCir is based on pretrained vision-language model and LLM model and it can generate pseudo triplet given only unlabeled images, it is easy to scale up to generate more data. **(b) There are still some failure cases, it is possible to enhance SynCir with stronger models.** Figure 5 shows some failure cases of our label synthesis method. In Figure 5(a), the target caption describes the target image with the location information "next to a forest", while the reference caption does not include this. LLM interprets this difference between two captions and generate the improper query text "Add a forest next to the train in the target image.". In Figure 5(b), the error is caused by LLM. LLM doesn't understand the difference between two captions and inaccurately interpret "hanging" as the text in the sign. To deal with these failure cases, stronger models can be adopted in the pipeline to generate better captions and query texts. Moreover, the data filter can be enhanced by considering more about the consistency between image and text. One possible solution is to use a well-trained CIR model to evaluate the quality of the synthetic triplet.

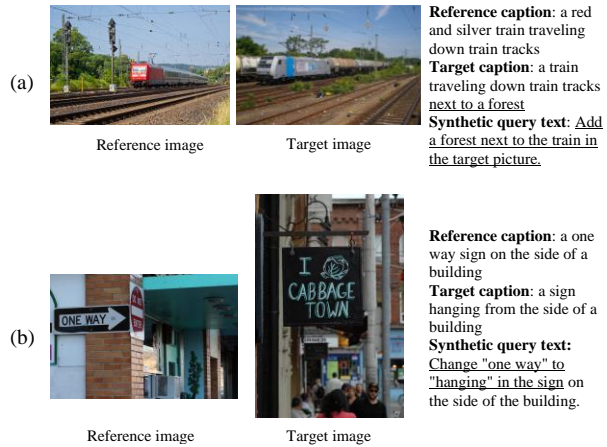

Figure 5: Some failure cases of generated labels. (a) The difference in reference caption and target caption leads to improper synthetic query text. (b) The misunderstanding of LLM leads to the inaccurate query text.

**Hybrid training for other ZS-CIR methods.** Our hybrid training strategy can be used with other ZS-CIR methods. Although current experiments are conducted on Pic2Word (Saito et al. 2023), it is similar to use it with other ZS-CIR methods. The approach is to add an additional contrastive loss for pseudo triplet supervision to ZS-CIR methods.

**Synthetic data for supervised methods**. Our synthetic data with CIR triplets can be used to train triplet-supervised methods, such as CLIP4CIR (Baldrati et al. 2022) and SPRC (Xu et al. 2024). Our experiment on hybrid training also verifies this. As shown in Table 5, training with only triplet supervision can achieve a relatively good performance on CIRR val set (average recall: 71.71%).

**Future work**. First, we will try to improve the quality of synthetic labels by using stronger pretrained models and enhanced data filter in the data synthesis pipeline. Second, we are interested in exploring the synthetic dataset working with triplet supervised methods. We want to know if the synthetic dataset can work with human-annotated CIR dataset to improve the performance of triplet supervised methods. Third, for ZS-CIR, the data distribution in the synthetic dataset is an important factor that influences the performance on different CIR datasets. We plan to generate synthetic labels for fashion images and include them in our synthetic dataset. We will verify its ZS-CIR performance on FashionIQ.

## Conclusion

In this paper, we try to improve zero-shot composed image retrieval (ZS-CIR) by using synthetic labels. We propose Hybrid CIR (HyCIR), which includes a CIR label synthesis pipeline SynCir to generate pseudo triplets and a hybrid training strategy to work with both ZS-CIR supervision and synthetic triplet supervision. SynCir pipeline consists of image pair extraction, vision-language model and LLM based query text generation, and language semantic based data filter. It can generate triplets from unlabeled images and makes it easy to scale up to more data. The hybrid training strategy makes existing ZS-CIR methods to work with the synthetic dataset generated by SynCir. Experiments with a Pic2Word (Saito et al. 2023) baseline show that we can improve ZS-CIR performance significantly. Our BLIP-based solution achieves SOTA performance on common CIR benchmarks: CIRR test set and CIRCO test set.